\documentclass{article}

\PassOptionsToPackage{numbers, sort&compress}{natbib}

    \usepackage[final]{neurips_distshift_2022}

\usepackage[T1]{fontenc}    
\usepackage[backref=page]{hyperref}       
\usepackage{url}            
\usepackage{booktabs}       
\usepackage{amsfonts}       
\usepackage{nicefrac}       
\usepackage{microtype}      
\usepackage{xcolor}         
\usepackage{graphicx}
\usepackage{amsmath,amssymb} 
\usepackage{pgfplotstable}
\usepackage[font=small]{caption}
\usepackage{subcaption}
\usepackage[rightcaption]{sidecap}
\usepackage[toc,page,header]{appendix}
\usepackage{minitoc}
\usepackage{todonotes}
\usepackage{tikz}
\usetikzlibrary{shapes}
\usepackage{ifthen}
\usepackage{siunitx}
\usepackage{collcell}
\usepackage{pgfplots}
\pgfplotsset{compat=newest} 
\usepgfplotslibrary{units} 
\usepackage{array}
\usepackage{adjustbox}

\usepackage[capitalize]{cleveref}
\crefname{section}{Sec.}{Secs.}
\Crefname{section}{Section}{Sections}
\Crefname{table}{Table}{Tables}
\crefname{table}{Tab.}{Tabs.}

\newcolumntype{R}[2]{%
    >{\adjustbox{angle=#1,lap=\width-(#2)}\bgroup}%
    l%
    <{\egroup}%
}
\newcommand*\rot{\multicolumn{1}{R{45}{1em}}}

\newcommand{\markerMLO}{\raisebox{0.5pt}{\tikz{\node[draw,scale=0.4,circle,fill=blue](){};}}}
\newcommand{\markerBLO}{\raisebox{0.5pt}{\tikz{\node[draw,scale=0.4,regular polygon, regular polygon sides=4,fill=red](){};}}}
\newcommand{\markerBLT}{\raisebox{0.5pt}{\tikz{\node[draw,scale=0.4,regular polygon, regular polygon sides=4,fill=violet!20!white](){};}}}
\newcommand{\markerMLT}{\raisebox{0.5pt}{\tikz{\node[draw,scale=0.3,regular polygon, regular polygon sides=3,fill=teal!10!teal,rotate=0](){};}}}

\title{A new benchmark for group distribution shifts in hand grasp regression for object manipulation. \\Can meta-learning raise the bar?}

\author{
    Théo Morales\\
    Trinity College Dublin\\
    Dublin, Ireland \\
    \texttt{moralest@tcd.ie}
    \and
	Gerard Lacey\\
    Maynooth University\\
    Maynooth, Ireland\\
    \texttt{gerard.lacey@.mu.ie}
}

\begin{document}
\maketitle
\doparttoc 
\faketableofcontents 

\begin{abstract}
Understanding hand-object pose with computer vision opens the door to new applications in mixed reality, assisted living
or human-robot interaction. Most methods are trained and evaluated on balanced datasets.
This is of limited use in real-world applications; how do these methods perform in the wild on unknown objects?
We propose a novel benchmark for object group distribution shifts in hand and object pose regression.
We then test the hypothesis that meta-learning a baseline pose regression neural network can adapt
to these shifts and generalize
better to unknown objects. Our results show measurable improvements over the baseline, depending on the amount
of prior knowledge. For the task of joint hand-object pose regression, we observe optimization interference for
the meta-learner. To address this issue and improve the method further, we provide a comprehensive analysis
which should serve as a basis for future work on this benchmark.
\end{abstract}

\section{Introduction}
\label{sec:intro}

Joint hand-object pose regression methods -- and hand pose regression in object grasping --
are not commonly benchmarked for in-the-wild data or a wide variety of object grasps \citep{HUANG2021207}.
They aim to generalize to unseen poses on the same objects by learning from a
large collection of diverse poses and grasps \citep{hampali2020honnotate,dexycb,FirstPersonAction_CVPR2018}.
This may be due to the cost of annotating 3D pose and the limited availability of 3D scanned objects, which
encourages researchers to reuse objects' meshes. In addition, synthetic datasets of realistic object grasps
are hard to produce and lead to domain adaptation challenges. However, such models have 
limited use if they are only accurate for a limited set of objects.

To accurately predict the pose of a hand occluded by an object, a deep learning model must learn prior
knowledge of various grasps of diverse objects.
Neural networks are usually trained to keep a balance between generalization and specialization
as their knowledge is frozen when deployed. However, temporarily learning at test time would
allow them to specialize their parameters for a specific object grasp while remaining
generalisable by forgetting this specialized knowledge.
Test-time adaptation (TTA) methods demonstrate improvements in regression accuracy for distribution shifts
and in-the-wild data in other domains, such as human pose estimation \citep{Hao2021TestTimePW},
gaze estimation \citep{Park2019FewShotAG}, mesh reconstruction \citep{Li2020OnlineAF} and video object
segmentation \citep{Xiao2020OnlineMA}.
We propose to achieve this on the grasp prediction problem with a meta-learning algorithm,
where the goal is to quickly learn new tasks from a few examples at test time \citep{Thrun1998LearningTL,Lake2016BuildingMT,andrychowicz_learning_2016}.
We then evaluate this method on a novel benchmark for group distribution shifts in hand-object pose regression
for object grasping.
How does the performance of a CNN pose predictor evolve as the test set grasps
diverge from the training set? We answer this question and look at the advantages and limitations of
meta-learning for this application via experiments and empirical analysis.

\paragraph{Contributions}
In this work, we: (a) reformulate grasp prediction in the context of multi-task learning such that
meta-learning can be applied, 
(b) propose a new benchmark for object group shifts in hand grasp regression based on the DexYCB
dataset \citep{dexycb}, and (c) prove that meta-learning is effective at tackling group distribution
shifts for hand grasp regression.
We also find an increase in accuracy for unknown objects from 6 training objects upwards.
We compare the relative error of the meta-learning with a baseline on our benchmark and provide a comprehensive analysis of the limitations of the method.

\section{A benchmark for object group shifts in hand grasp regression}

In this section, we explain the task set creation process which is necessary so
that meta-learning can be applied to the grasp prediction problem. 

\paragraph{Task set creation}\label{sec:procedure}

To cast pose prediction as a multi-task problem and apply meta-learning, we must create a
dataset of tasks from a dataset of samples. For such a task set, we require that:

\begin{itemize}
	\item Each task is composed of a support set of $K$ randomly sampled images corresponding to
		one manipulation sequence of an object by one subject,
		and a query set of $Q$ distinct random samples from the said sequence.
	\item A series of $\Omega$-objects-left-out splits is used, where $\Omega$ is the number of
    	objects absent in the training split and placed in the test split.
		Thus, all images associated with 
		$\Omega+\min(5, \frac{\Omega}{2}+(\Omega\mod2))$ objects are removed from the training split: the samples
		associated with $5$ or fewer objects are used for validation while the
		ones for other $\Omega$ objects are used for testing. This ensures that there is no overlap
		between the training, validation and test splits.
	\item For all values of $\Omega$, the objects are randomly sampled with a fixed random seed
	for reproducibility across experiments.
\end{itemize}

We use the DexYCB dataset to create our task set and run our experiments, although the procedure is
applicable to any object manipulation dataset (see the survey of \citep{HUANG2021207} for an overview of hand-object
pose datasets). Along with a similarity study
of object grasps from this dataset, we provide more implementation details as well as
visual examples of tasks in \cref{appendix:dataset}, and the code on GitHub 
\footnote{https://github.com/DubiousCactus/meta-learning-HOPE}.

\section{Evaluating the effectiveness of meta-learning}
In this section, we describe our experimental framework used to assess the hypothesis that TTA of a pose prediction
model through meta-learning improves the generalization of unknown objects.
We give an introduction to optimization-based meta-learning in \cref{appendix:metalearning}.

\subsection{Experimental framework}

\begin{figure}[h]
    \centering
    \includegraphics[width=0.6\textwidth]{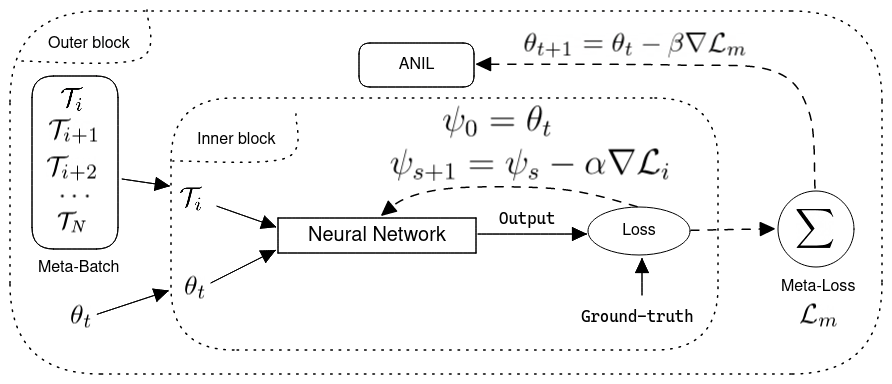}
	\caption{\textbf{An overview of the meta-learning system.} -- The outer block optimizes the
	inner block's parameters by back-propagating the second-order gradients of the meta-loss. This
cost function is the cumulative loss of the inner block on all $N$ tasks, computed with the adapted
parameters $\psi_S$ obtained after $S$ optimization steps and with the same initial parameters $\theta_t$.}
    \label{fig:metalearning-diagram}
\end{figure}

To evaluate the effectiveness of meta-learning at tackling object group shifts and dealing with highly imbalanced
dataset splits, we apply ANIL \citep{ANIL} with the improvements of MAML++ \citep{Antoniou} to a ResNet18 \citep{resnet} baseline.
As seen in \cref{fig:metalearning-diagram}, our framework consists of an outer block (ANIL) and an inner block (ResNet18).


\paragraph{Outer block}
The outer block trains the inner block to learn good parameters that are amenable to fast adaptation
with few gradient descent steps and examples.
It is based on MAML and is a re-implementation of ANIL \cite{ANIL} which brings performance
improvements to the former by only adapting the head of the network and learning a common feature
extractor for all tasks. We further implement some of the improvements proposed by \cite{Antoniou},
such as the Multi-Step Loss 
and Derivative-Order Annealing. 
Furthermore, to deal with the meta-overfitting phenomenon
that arises from the non-mutual-exclusiveness of our regression tasks, we implement the regularization
method of \cite{Yin2020MetaLearningWM}. We use the \textit{learn2learn} \cite{l2l} library to facilitate
the implementation and minimize mistakes.

\paragraph{Inner block}
For the inner block, the baseline pose regression neural network, we use the same backbone for both
experiments: ResNet18 \cite{resnet} pre-trained on ImageNet \cite{imagenet}. The choice of a shallow
ResNet architecture is motivated by the speed of training and low memory footprint for meta-learning.
It allows a fair comparison without being impacted by the limitations of meta-learning methods
regarding more complex architectures.
Combining these more complex and efficient architectures with meta-learning should be
the focus of future work. For a fair comparison, the baseline is trained with weight decay regularization
to minimize overfitting.


For the baseline, we use a batch size of $64$ and a learning rate $\alpha = 10^{-3}$ for $100$
epochs. For the meta-learner, we use a meta-batch size of $8$ and found learning rates
$\alpha = 10^{-5}$ and $\beta = 10^{-2}$ via hyper-parameter search using \textit{Weights \& Biases} \citep{wandb},
with $300$ epochs.
For both methods,  we normalize the inputs according to ImageNet's statistics 
and align the ground-truth poses to the root of the hand.
Using wrist-aligned pose labels greatly improves the convergence of both methods,
and the absolute position in 3D space is not in the scope of this work. 
For the meta-learner, we use $K=10$ for the support set, $Q=50$ for the query set, and $15$ adaptation
steps. This implies that during evaluation, there are $K \times
(\frac{T}{K+Q})$ samples -- with $T$ being the size of the test split -- virtually removed from the
test split, which is a slight disadvantage for the baseline. It is acceptable because the goal is
to measure error changes relative to various test splits. In fact, the pose prediction error of the 
meta-learner is roughly $1.5\times$ that of the baseline for both experiments.
When evaluating the meta-learner, we average all metrics over $5$ runs to account for the run-time
stochasticity coming from the support and query sets sampling.
For all experiments, we normalize the error against each model's
performance on the easiest setting (5 test objects) to compare the relative error changes.

With this framework, we design two experiments:
(1) we evaluate the relative error of the 3D hand pose regression for manipulation samples
of unknown objects, (2) we reiterate on the joint hand-object pose regression by incorporating
the object bounding cuboid coordinates in the targets, as
in \citep{Doosti2020HOPENetAG,Li2021ArtiBoostBA,huang_hot-net_2020}.
\subsection{Results}
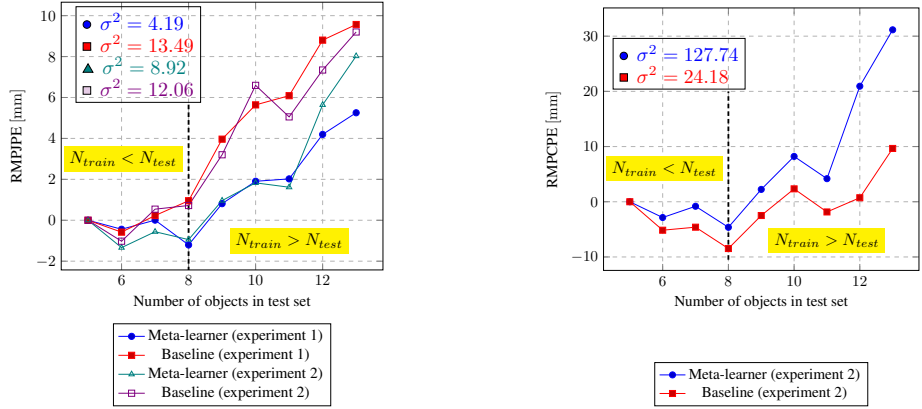
\begin{figure}[tb]
    \centering
     \begin{subfigure}[t]{0.49\textwidth}
         \begin{center}
         \resizebox{0.75\textwidth}{!}{
            \begin{tikzpicture}
              \begin{axis}[
                  width=1.3\linewidth, 
                  grid=major, 
                  grid style={dashed,gray!60}, 
                  xlabel=Number of objects in test set, 
                  ylabel=RMPJPE,
                  y unit=\si{mm},
                  legend style={at={(0.5,-0.2)},anchor=north}, 
                ]
                \addplot table[x=held out,y=mpjpe,col sep=comma] {tab/handonly/anil_mpjpes_origin.csv}; 
                \addplot table[x=held out,y=mpjpe,col sep=comma] {tab/handonly/baseline_mpjpes_origin.csv}; 
                \draw [densely dashed,black,very thick] (8,-2.5) -- (8,5.8);
                \node[fill=yellow,text=black,align=left,scale=1.1] at (6.05,3) {$N_{train} < N_{test}$};
                \node[fill=yellow,text=black,align=left,scale=1.1] at (11,-1) {$N_{train} > N_{test}$};
                \node[draw,align=left,scale=1.3] at (6.5,8.1) {\markerMLO \hspace{2px} \color{blue}{$\sigma^2=4.19$} \\
                \markerBLO \hspace{2px} \color{red}{$\sigma^2 = 13.49$} \\
                \markerMLT \hspace{2px} \color{teal}{$\sigma^2 = 8.92$} \\
                \markerBLT \hspace{2px} \color{violet}{$\sigma^2 = 12.06$}};
                \addplot+[teal,mark=triangle] table[x=held out,y=mpjpe,col sep=comma] {tab/handobj/anil_mpjpes_origin.csv}; 
                \addplot+[violet,mark=square] table[x=held out,y=mpjpe,col sep=comma] {tab/handobj/baseline_mpjpes_origin.csv}; 
                
                \legend{Meta-learner (experiment 1), Baseline (experiment 1), Meta-learner (experiment 2), Baseline (experiment 2)}
              \end{axis}
            \end{tikzpicture}
            }
            \caption{\textbf{Hand only} -- For both experiments, the error increases less steeply and has lower variance $\sigma^2$
            as the number of test objects grows for the meta-learner. The latter is better at adapting to dataset imbalance.}
            \label{fig:mpjpe}
        \end{center}
     \end{subfigure}
     \hfill
     \begin{subfigure}[t]{0.49\textwidth}
      \begin{center}
         \resizebox{0.75\textwidth}{!}{
        \begin{tikzpicture}
          \begin{axis}[
              width=1.3\linewidth, 
              grid=major, 
              grid style={dashed,gray!60}, 
              xlabel=Number of objects in test set, 
              ylabel=RMPCPE,
              y unit=\si{mm},
              legend style={at={(0.5,-0.35)},anchor=north}, 
            ]
            \addplot table[x=held out,y=mpcpe,col sep=comma] {tab/handobj/anil_mpcpes_origin.csv}; 
            \addplot table[x=held out,y=mpcpe,col sep=comma] {tab/handobj/baseline_mpcpes_origin.csv}; 
            \draw [densely dashed,black,very thick] (8,-10.5) -- (8,20);
            \node[fill=yellow,text=black,align=left,scale=1.1] at (6.05,6) {$N_{train} < N_{test}$};
            \node[fill=yellow,text=black,align=left,scale=1.1] at (11,-7) {$N_{train} > N_{test}$};
            \node[draw,align=left,scale=1.3] at (6.5,25) {\markerMLO \hspace{2px} \color{blue}{$\sigma^2 = 127.74$}\\
            \markerBLO \hspace{2px} \color{red}{$\sigma^2 = 24.18$}};       
            \legend{Meta-learner (experiment 2), Baseline (experiment 2)}
          \end{axis}
        \end{tikzpicture}
        }
        \caption{\textbf{Object only} -- Both methods exhibit a high variance but the meta-learner is worse by an order of magnitude.
        The latter isn't able to reduce the object pose error.}
        \label{fig:mpcpe}
        \end{center}
     \end{subfigure}
     \caption{\textbf{Relative Mean Per-Joint Pose Error (RMPJPE) \& Mean Per-Corner Pose Error (RMPCPE) as functions of the imbalance level}
     -- All curves are aligned to the origin to show the relative changes and compare the generalisability of both methods.
     The more objects are in the test set, the more deprived the training set (see \cref{tab:samples_per_split} in \cref{appendix:dataset});
     the dashed line shows the equilibrium between the two.
     Both experiments are plotted in \cref{fig:mpjpe} since they both regress the hand pose, while only experiment 2 predicts the object pose
     and is plotted in \cref{fig:mpcpe} on its own.
     The meta-learner is overall better at handling object class imbalance, and thus generalisability to unknown grasps. However, this behaviour
     is not seen for the prediction of the object corners coordinates.}
     \label{fig:mpjpe_mpcpe}
\end{figure}

\paragraph{Experiment 1: hand pose only}\label{sec:expone}
\cref{fig:mpjpe} shows the meta-learner's ability to deal with object group shifts on a macro scale.
The size of the training set decreases inversely and in proportion to the size of the test set,
thus the error is expected to go up for the baseline as the test set grows.
The error of both models rises steeply from 8 objects in the test set, where the gap starts to widen,
but less steeply for the meta-learner.
It corresponds to 8 training objects as well (with 4 for validation); a more or less balanced dataset.
By fitting a linear regression model on both curves, we prove that the difference in the two slopes
is statistically significant with a $p$-value of $0.0031$.

We look at the micro scale by freezing the training split for 3 levels of prior knowledge, and
progressively adding objects to the test split; the results are shown in
\cref{fig:mpjpe_fine}.
Both methods are trained and evaluated on 3 train/test splits; the average curve is shown for the 3 sizes.
The meta-learner and the baseline behave
identically in \cref{fig:mpjpe_fine_3} since the training set is too small for the
meta-learner to collect sufficient prior knowledge for adaptation: 3 training objects are not
sufficient prior knowledge. As the training set size and
variability grow for \cref{fig:mpjpe_fine_6}, the meta-learner reduces the error significantly
better than the baseline: 6 objects are enough prior knowledge.
However, with 9 training objects in \cref{fig:mpjpe_fine_9}, both curves have a similar horizontal 
slope and a variance roughly a third of that of \cref{fig:mpjpe_fine_3}.
In short, the baseline and the meta-learner generalize equally well.
For this case, the improvement of the meta-learner may nonetheless be revealed with more test objects
if available. The findings are different than in \cref{fig:mpjpe} since the models are here compared in terms of accuracy
for a given training set size, and not their ability to deal with dataset imbalance.

\input{tab/error_curves_individual_mean}

\paragraph{Experiment 2: joint hand-object pose}\label{sec:exptwo}
Analogously to the previous experiment, \cref{fig:mpjpe} shows a mild improvement in the hand pose regression for the
meta-learner when trained for the task of joint hand-object pose regression. This improvement
however is lesser than for the first experiment, as reflected by the smaller gap between the two
curves and a $p$-value of $0.3186$. The null hypothesis cannot be rejected: the meta-learner shows
no significant improvement in generalization. \cref{fig:mpcpe} confirms this phenomenon and shows 
even worse results than the baseline. We tried increasing the network's capacity but obtained similar results with underfitting.
Therefore we can hypothesize that this joint regression causes interference in the adaptation to both tasks, and this problem
would have to be processed separately in future work. In \cref{sec:analysis}, we run more experiments to assess the hypotheses
made on the meta-learner in an attempt to explain the results.
\section{Empirical analysis and conclusions}\label{sec:analysis-conclusions}
\paragraph{Takeaways} 
From these experiments, we can conclude three things:
(1) for hand grasp prediction, the meta-learner is better able to deal with dataset imbalance than the baseline,
as shown in \cref{fig:mpjpe} this is especially true when using less than 8 training objects (or more than 8 test objects),
(2) the meta-learner improves the accuracy with enough training data (i.e. 6 objects) on unknown objects, and
(3) this method is ineffective for joint hand-object pose regression,
and is even worse than the baseline for object pose prediction (see \cref{fig:mpcpe}). This weakness is investigated
in the following section.

\subsection{Empirical analysis}\label{sec:analysis}
\paragraph{Hypothesis 1: the model learns specialized object-specific parameters}
ANIL and the like learn prior knowledge of related tasks during training so that they can adapt 
to specialized parameters for a novel task with few optimization steps \citep{Huisman2021ASO}. 
But does it translate to learning object- or grasp-specific parameters in the proposed framework?
To verify this, we plotted the t-SNE embeddings of the network's head parameters post adaptation (see \cref{appendix:tsne}).
We would expect to find clusters of tasks for the same manipulated object but none appeared.
It reveals that the parameters do not specialize to a specific grasp or object after adaptation, thus refuting this hypothesis.
This may well be due to the \textit{non-mutual exclusiveness} of tasks incurring memorization overfitting 
\citep{rajendran_meta-learning_2020}, for which the regulariser of \citep{Yin2020MetaLearningWM} did not help.
In future work, the problem should be properly formulated in the meta-objective to encourage specialization.

\paragraph{Hypothesis 2: Using Oriented Bounding Box (OBB) coordinates in the training signal constrains the hand-object pose}
In experiment 2, we rely on this hypothesis and make the conjecture that 
OBBs differ enough to lead away from the initialization in the loss landscape. In truth, \cref{fig:mpcpe} shows
that it limits the adaptation for the hand pose and leads to worse generalization than the baseline for the object pose.
\citep{Dryden1998StatisticalSA} define the shape as \textit{"all the geometrical information that
remains when location, scale and rotational effects are filtered out from an object"}. Therefore we
can consider that most OBBs have almost identical shapes, except for unusually thin or wide objects
such as \textit{large marker} or \textit{pitcher base}. We postulate that due to this, the adaptation
phase should produce small gradients for the object keypoints and thus the hand keypoint regression should
not be severely impacted. However, we observed a $2-3\times$ increase in gradient norms from experiment 1 to 2 
(see \cref{appendix:gradnorm} for more details).
We hypothesize that this is due to the complex hand-object relationship 
which is not expressed in the meta-objective, causing interference in the meta-optimization landscape.
In future work, these constraints should be explicitly defined in the objective, or the two problems decoupled.

\paragraph{Conclusions.}
Our results show measurable improvements over the baseline, where it reduces the error rise with less
than 8 training objects. On the other hand, it can generalize better than the baseline from 6 training objects.
For the task of joint hand-object pose regression, the meta-learner's ability to deal with object group shifts
is dampened and is worse than the baseline for the object pose. In our empirical analysis, we show that this
may be due to interference in the optimization. We further propose solutions to address this issue and to
encourage the model to learn object-specific parameters during adaptation. This should in turn improve the
effectiveness of this method.

\begin{ack}
This work was conducted with the financial support of the Science
Foundation Ireland Centre for Research Training in Digitally-Enhanced Reality (d-real) under Grant
No. 18/CRT/6224. For the purpose of Open Access, the author has applied a CC BY public copyright
licence to any Author Accepted Manuscript version arising from this submission.
\end{ack}

\bibliographystyle{plainnat}
\bibliography{main}


\appendix
\addcontentsline{toc}{section}{Supplementary material} 
\part{Appendix} 
\parttoc 



\section{Task set creation}
\label{appendix:dataset}
\begin{figure}[tb]
    \centering
    \includegraphics[width=\textwidth]{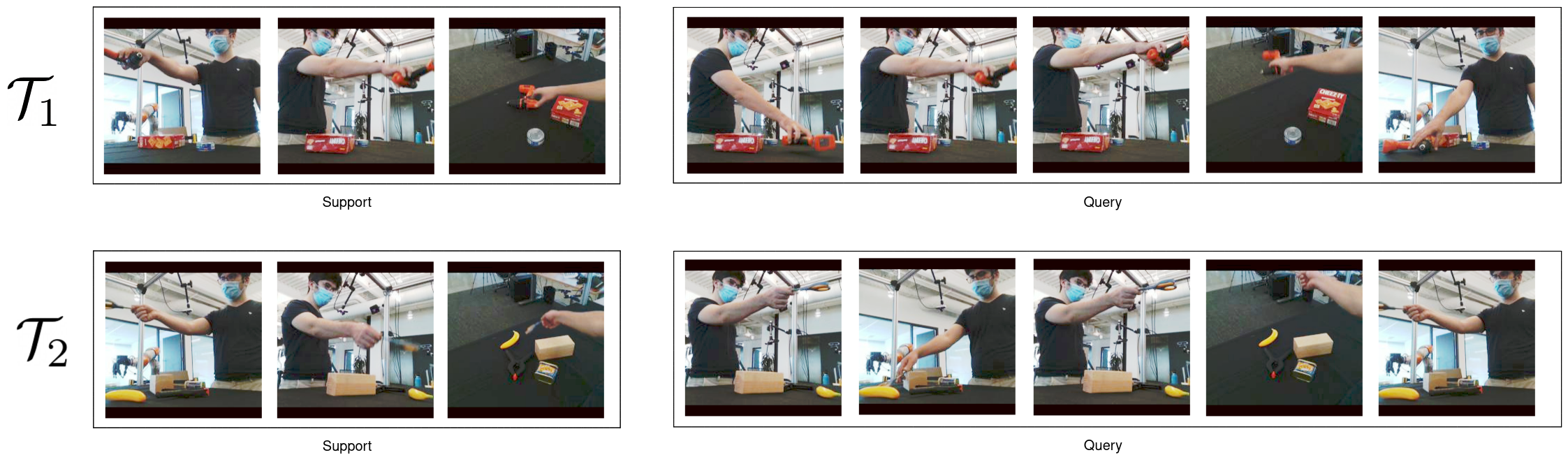}
    \caption{\textbf{Example of tasks built from the DexYCB dataset.} -- 
	$\mathcal{T}_1$ and $\mathcal{T}_2$ are two tasks composed of a support set of 3 training images,
	and a query set of 5 test images for 3-shot learning. $\mathcal{T}_1$ is the manipulation of a
	drill, while $\mathcal{T}_2$ is the grasping of scissors. The tasks contain several viewpoints but
	only one subject and intent of use in their support and query sets.} \label{fig:dexycb-tasks}
\end{figure}

\begin{table}[h]
    \centering
    \caption{
        \textbf{Image samples per split as the number of unseen objects grows in the test split} -- As the test set size increases for each added object, the
        training set size decreases proportionally. We cap the validation objects so as to maximise the number of dataset versions.\\
    } 
    \resizebox{\linewidth}{!}{ 
        \begin{tabular}{@{}lccccccccc@{}}
            \toprule
            \# objects in test split & 5 & 6 & 7 & 8 & 9 & 10 & 11 & 12 & 13 \\
            \midrule
            Training    & $242,704$ & $221,688$   & $183,104$ & $163,608$   & $117,824$ & $98,704$    & $80,072$    & $58,016$  & $42,288$ \\
            Validation  & $60,248$  & $62,552$    & $79,232$  & $82,240$    & $105,040$ & $101,744$   & $99,952$    & $100,896$ & $99,072$ \\
            Test        & $100,960$ & $119,672$   & $141,576$ & $158,064$   & $181,048$ & $203,464$   & $223,888$   & $245,000$ & $262,552$ \\
            
            \bottomrule
        \end{tabular}
    } 
    \label{tab:samples_per_split}
\end{table}
In order to assess the generalisation of a grasp prediction model to unseen objects, we need a
dataset of 2D images of the largest amount of manipulated objects with full 3D hand
joint annotations, as well as 6D object pose. The review of \citep{HUANG2021207} provides an
extensive list of available datasets, but their limited amount of objects was the main factor in
choosing the unlisted and recently released DexYCB \citep{dexycb}.
We began the experiment with the HO-3Dv3 dataset \citep{hampali2020honnotate, Hampali2021HO3Dv3IT}
which contains 10 different objects. This limitation hindered the experimental results, as more
objects are needed to have enough training, validation and testing object categories, since they
cannot overlap in each split. We further used the DexYCB after realising that the results were
not conclusive.
It contains $582,000$ annotated frames with $10$ subjects and $21$ objects, of which the
\textit{large clamp} is not annotated and therefore removed.
We discard all frames where the hand is not in contact with the object, such that at least
two fingertips are within the object bounding box. 
That oriented cuboid is computed using Trimesh \citep{trimesh} on the associated object mesh,
before applying transforms.
\cref{tab:samples_per_split} shows the final size of each split for every variant of the dataset,
and \cref{fig:dexycb-tasks} shows examples of tasks built from DexYCB.

With the DexYCB dataset, there are approximately 20K samples per object with 100 manipulation sequences each,
providing the across-task diversity required for the training of neural networks and minimise overfitting.
Meta-learning shows its power over standard gradient descent optimisation in the presence of across-task diversity,
such that each task is an objective requiring specialisation.
We analyse the diversity of object-specific grasps using the Generalised Procrustes Analysis and the Procrustes distance
to compare mean hand shape similarity.
It is defined as the sum of the squared vertex distances:
\begin{equation}
d = \sum_i^{N} [(x_{i1} - x_{i2})^2+(y_{i1} - y_{i2})^2+(z_{i1} - z_{i2})^2]
\end{equation}
for two 3D shapes with $N$ matching vertices. The heat map of hand poses similarity in \cref{tab:dist_mat}
shows relatively few light cells, hence most grasps are similar to each other. This is partly because subjects may grasp
the same object in various and unusual ways, thus resulting in an uninformative mean hand shape. Due to this, we construct tasks from
individual manipulation sequences such that there is only one grasp per task, as described in \cref{sec:procedure}.
\begingroup

\renewcommand*\rot[2]{\multicolumn{1}{R{#1}{#2}}}

\newcommand*{\MinNumber}{0}
\newcommand*{\MaxNumber}{0.1455}
\newcommand{\ApplyGradient}[1]{
        \pgfmathsetmacro{\PercentColor}{100.0*(#1-\MinNumber)/(\MaxNumber-\MinNumber)}
        \hspace{-0.33em}\colorbox{yellow!\PercentColor!red}{}
}
\newcolumntype{G}{>{\collectcell\ApplyGradient}c<{\endcollectcell}}
\renewcommand{\arraystretch}{0}
\setlength{\fboxsep}{3mm} 
\setlength{\tabcolsep}{0pt}

\begin{table}[h]
    \centering
    \caption{
        \textbf{Distance heat map of mean hand shapes} -- Heat map of Procrustes distances of mean hand shapes obtained via
        Generalised Procrustes Analysis; similarity ranges from yellow to red (best seen in colour).
        The \textit{Large marker} induces the most different grasps in the dataset, while all box-shaped objects induce similar grasps (as seen by the more uniform upper left square).
        Note that all objects have several grasps associated with them, which vary depending on the intent of use of the subject.\\
    }\label{tab:dist_mat}
        \resizebox{0.65\linewidth}{!}{ 
            
            \begin{tabular}{r*{20}{G}}
                &
                \rot{45}{1em}{Master chef can} &
                \rot{45}{1em}{Cracker box} &
                \rot{45}{1em}{Sugar box} &
                \rot{45}{1em}{Tomato soup can} &
                \rot{45}{1em}{Mustard bottle} &
                \rot{45}{1em}{Tuna fish can} &
                \rot{45}{1em}{Pudding box} &
                \rot{45}{1em}{Gelatin box} &
                \rot{45}{1em}{Potted meat can} &
                \rot{45}{1em}{Banana} &
                \rot{45}{1em}{Pitcher base} &
                \rot{45}{1em}{Bleach cleanser} &
                \rot{45}{1em}{Bowl} &
                \rot{45}{1em}{Mug} &
                \rot{45}{1em}{Power drill} &
                \rot{45}{1em}{Wood block} &
                \rot{45}{1em}{Scissors} &
                \rot{45}{1em}{Large marker} &
                \rot{45}{1em}{Extra large clamp} &
                \rot{45}{1em}{Foam brick} \\
                Master chef can \hspace{0.33em}			& 0.0000 & 0.0061 & 0.0226 & 0.0183 & 0.0237 & 0.0300 & 0.0309 & 0.0533 & 0.0220 & 0.0797 & 0.0249 & 0.0238 & 0.0873 & 0.0454 & 0.0740 & 0.0019 & 0.1053 & 0.1455 & 0.0766 & 0.0423 \\ 
                Cracker box \hspace{0.33em}				& 0.0061 & 0.0000 & 0.0064 & 0.0057 & 0.0080 & 0.0130 & 0.0131 & 0.0301 & 0.0090 & 0.0486 & 0.0138 & 0.0071 & 0.0538 & 0.0228 & 0.0437 & 0.0035 & 0.0680 & 0.1016 & 0.0487 & 0.0210 \\ 
                Sugar box \hspace{0.33em}				& 0.0226 & 0.0064 & 0.0000 & 0.0030 & 0.0011 & 0.0029 & 0.0027 & 0.0106 & 0.0025 & 0.0216 & 0.0087 & 0.0008 & 0.0304 & 0.0062 & 0.0195 & 0.0148 & 0.0339 & 0.0590 & 0.0227 & 0.0056 \\
                Tomato soup can \hspace{0.33em}			& 0.0183 & 0.0057 & 0.0030 & 0.0000 & 0.0022 & 0.0034 & 0.0049 & 0.0138 & 0.0020 & 0.0299 & 0.0148 & 0.0028 & 0.0423 & 0.0096 & 0.0252 & 0.0112 & 0.0438 & 0.0713 & 0.0320 & 0.0077 \\ 
                Mustard bottle \hspace{0.33em}			& 0.0237 & 0.0080 & 0.0011 & 0.0022 & 0.0000 & 0.0028 & 0.0039 & 0.0096 & 0.0013 & 0.0211 & 0.0113 & 0.0007 & 0.0349 & 0.0060 & 0.0167 & 0.0154 & 0.0342 & 0.0583 & 0.0237 & 0.0041 \\ 
                Tuna fish can \hspace{0.33em}			& 0.0300 & 0.0130 & 0.0029 & 0.0034 & 0.0028 & 0.0000 & 0.0006 & 0.0043 & 0.0016 & 0.0157 & 0.0116 & 0.0043 & 0.0309 & 0.0029 & 0.0173 & 0.0206 & 0.0263 & 0.0482 & 0.0160 & 0.0024 \\ 
                Pudding box \hspace{0.33em}				& 0.0309 & 0.0131 & 0.0027 & 0.0049 & 0.0039 & 0.0006 & 0.0000 & 0.0047 & 0.0029 & 0.0147 & 0.0097 & 0.0049 & 0.0254 & 0.0028 & 0.0180 & 0.0215 & 0.0254 & 0.0469 & 0.0145 & 0.0031 \\ 
                Gelatin box \hspace{0.33em}				& 0.0533 & 0.0301 & 0.0106 & 0.0138 & 0.0096 & 0.0043 & 0.0047 & 0.0000 & 0.0079 & 0.0056 & 0.0170 & 0.0125 & 0.0222 & 0.0016 & 0.0101 & 0.0394 & 0.0128 & 0.0268 & 0.0060 & 0.0017 \\ 
                Potted meat can \hspace{0.33em}			& 0.0220 & 0.0090 & 0.0025 & 0.0020 & 0.0013 & 0.0016 & 0.0029 & 0.0079 & 0.0000 & 0.0220 & 0.0109 & 0.0032 & 0.0374 & 0.0057 & 0.0204 & 0.0140 & 0.0357 & 0.0599 & 0.0230 & 0.0040 \\ 
                Banana \hspace{0.33em}					& 0.0797 & 0.0486 & 0.0216 & 0.0299 & 0.0211 & 0.0157 & 0.0147 & 0.0056 & 0.0220 & 0.0000 & 0.0260 & 0.0235 & 0.0153 & 0.0072 & 0.0065 & 0.0624 & 0.0053 & 0.0115 & 0.0027 & 0.0086 \\ 
                Pitcher base \hspace{0.33em}			& 0.0249 & 0.0138 & 0.0087 & 0.0148 & 0.0113 & 0.0116 & 0.0097 & 0.0170 & 0.0109 & 0.0260 & 0.0000 & 0.0118 & 0.0324 & 0.0130 & 0.0267 & 0.0166 & 0.0398 & 0.0637 & 0.0206 & 0.0139 \\ 
                Bleach cleanser \hspace{0.33em}			& 0.0238 & 0.0071 & 0.0008 & 0.0028 & 0.0007 & 0.0043 & 0.0049 & 0.0125 & 0.0032 & 0.0235 & 0.0118 & 0.0000 & 0.0352 & 0.0073 & 0.0177 & 0.0161 & 0.0352 & 0.0607 & 0.0263 & 0.0060 \\ 
                Bowl \hspace{0.33em}					& 0.0873 & 0.0538 & 0.0304 & 0.0423 & 0.0349 & 0.0309 & 0.0254 & 0.0222 & 0.0374 & 0.0153 & 0.0324 & 0.0352 & 0.0000 & 0.0204 & 0.0240 & 0.0700 & 0.0247 & 0.0315 & 0.0174 & 0.0243 \\ 
                Mug \hspace{0.33em}						& 0.0454 & 0.0228 & 0.0062 & 0.0096 & 0.0060 & 0.0029 & 0.0028 & 0.0016 & 0.0057 & 0.0072 & 0.0130 & 0.0073 & 0.0204 & 0.0000 & 0.0088 & 0.0327 & 0.0143 & 0.0308 & 0.0082 & 0.0011 \\ 
                Power drill \hspace{0.33em}				& 0.0740 & 0.0437 & 0.0195 & 0.0252 & 0.0167 & 0.0173 & 0.0180 & 0.0101 & 0.0204 & 0.0065 & 0.0267 & 0.0177 & 0.0240 & 0.0088 & 0.0000 & 0.0565 & 0.0104 & 0.0200 & 0.0120 & 0.0084 \\ 
                Wood block \hspace{0.33em}				& 0.0019 & 0.0035 & 0.0148 & 0.0112 & 0.0154 & 0.0206 & 0.0215 & 0.0394 & 0.0140 & 0.0624 & 0.0166 & 0.0161 & 0.0700 & 0.0327 & 0.0565 & 0.0000 & 0.0856 & 0.1217 & 0.0599 & 0.0301 \\ 
                Scissors \hspace{0.33em}				& 0.1053 & 0.0680 & 0.0339 & 0.0438 & 0.0342 & 0.0263 & 0.0254 & 0.0128 & 0.0357 & 0.0053 & 0.0398 & 0.0352 & 0.0247 & 0.0143 & 0.0104 & 0.0856 & 0.0000 & 0.0047 & 0.0062 & 0.0181 \\ 
                Large marker \hspace{0.33em}			& 0.1455 & 0.1016 & 0.0590 & 0.0713 & 0.0583 & 0.0482 & 0.0469 & 0.0268 & 0.0599 & 0.0115 & 0.0637 & 0.0607 & 0.0315 & 0.0308 & 0.0200 & 0.1217 & 0.0047 & 0.0000 & 0.0141 & 0.0353 \\ 
                Extra large clamp \hspace{0.33em}		& 0.0766 & 0.0487 & 0.0227 & 0.0320 & 0.0237 & 0.0160 & 0.0145 & 0.0060 & 0.0230 & 0.0027 & 0.0206 & 0.0263 & 0.0174 & 0.0082 & 0.0120 & 0.0599 & 0.0062 & 0.0141 & 0.0000 & 0.0111 \\ 
                Foam brick \hspace{0.33em}				& 0.0423 & 0.0210 & 0.0056 & 0.0077 & 0.0041 & 0.0024 & 0.0031 & 0.0017 & 0.0040 & 0.0086 & 0.0139 & 0.0060 & 0.0243 & 0.0011 & 0.0084 & 0.0301 & 0.0181 & 0.0353 & 0.0111 & 0.0000 \\ 
            \end{tabular}
        
        } 
    
\end{table}

\endgroup

\section{Optimisation-based meta-learning}
\label{appendix:metalearning}

A task is defined as $\mathcal{T}_i = \{p_i(\mathbf{x}), p_i(\mathbf{y}|\mathbf{x}),
\mathcal{L}_i\}$ and comprises a distribution of samples $p_i(\mathbf{x})$, a distribution of
ground-truth labels $p_i(\mathbf{y}|\mathbf{x})$ and a loss function $\mathcal{L}_i$. By targeting
similar tasks, a model can make use of the shared structure in the data across tasks for a
meta-learning approach. This allows the model to learn a generic -- or across-task -- prior, while
having the plasticity required to adjust its parameters for task-specific knowledge.

Meta-learning is often presented in the context of few-shot learning (also known as $K$-shot
learning), where the goal is to optimise a new objective from a few examples, referred to as the
context set. During training, the meta-learner optimises the learner on various objectives
from their context set (of $K \leq 10$ samples typically) and evaluates its performance on a
validation set of $Q$ unseen samples called the query set. The model's performance on the 
tasks aggregate of the latter is a measure of its ability to learn quickly.

In a single-task supervised learning context, we consider a single dataset
$\mathcal{D}=\{(\mathbf{x},\mathbf{y})_K\}$ of pairs of images and labels. The optimisation problem
is thus to minimise the loss over the entire dataset to find the optimal model parameters $\theta$ as:

\begin{equation}\label{eq:typical}
    \min_\theta \mathcal{L}(\theta, \mathcal{D}),
\end{equation}

where $\mathcal{L}$ is the \textit{Mean Squared Error} for regression tasks. It is minimised
such that the update rule at each optimisation step is given by
\begin{equation}\label{eq:typical_update}
    \theta_{t+1} = \theta_t - \alpha_t \nabla f_{\theta_t}
\end{equation}
with a static or adaptive learning rate $\alpha$ for each step $t$ and the optimised function
$f$ parameterised by $\theta_t$ at a step $t$. The most characteristic aspect of the meta-learning
approach is that the update rule is instead learnt end-to-end, such that the parameter update can be reformulated,
as defined by \citep{andrychowicz_learning_2016}, to
\begin{equation}\label{eq:meta_update}
    \theta_{t+1} = \theta_t + g_t(\nabla f_{\theta_t}, \phi)   
\end{equation}
where $g$ is the learnt optimiser parameterised by $\phi$ and $f$ is the optimisee.

Typically, an optimisation-based meta-learning system is composed of an outer block called the
optimiser, which learns the update rule of the base learner, and of an inner block called the
optimisee, which is the base learner that directly learns the task. The optimisee is thus optimised
by the optimiser block to rapidly learn novel tasks. 
A general definition of the optimisation problem for meta-learning is given as
\begin{equation}\label{eq:maml_concise}
     \min_\theta \sum_{i=1}^N \mathcal{L}_i(\theta, \mathcal{D}_i) 
\end{equation}
for $N$ tasks where each task is a small dataset $\mathcal{D}_i$. For MAML and other optimisation-based
meta-learning algorithms, this meta-objective becomes
\begin{equation}\label{eq:maml}
    \min_\theta \sum_{\mathcal{T}_i \sim p(\mathcal{T})}^N \mathcal{L}_{\mathcal{T}_i}(f_{\psi}) =
    \sum_{\mathcal{T}_i \sim p(\mathcal{T})}^N \mathcal{L}_{\mathcal{T}_i}(f_{\theta - \alpha \nabla_\theta \mathcal{L}_{\mathcal{T}_i}(f_\theta)})
\end{equation}
for $N$ tasks sampled from $p(\mathcal{T})$, with post-adaptation parameters $\psi$.
As for the update rule, it is defined as such:
\begin{equation}\label{eq:maml_update}
    \theta_{t+1} = \theta_t - \beta \nabla_{\theta_t} \sum_{\mathcal{T}_i \sim p(\mathcal{T})}^{N}
    f_{\theta_t - \alpha\nabla_{\theta_t} \mathcal{L}_{\mathcal{T}_i}(f_{\theta_t})}   \text{.}
\end{equation}

In most cases, the learnt optimiser is only effective during training. However, some algorithms propose to learn per-layer
and/or per-step learning rates for SGD as the learnable optimiser, and use them at test-time during the adaptation
phase \citep{Antoniou}.

\section{Empirical analysis}\label{appendix:analysis}
In this section, we run extra experiments on the trained models to support the analysis of the hypotheses
made in \cref{sec:analysis}.

\subsection{Visualisation of the specialised networks' parameters}\label{appendix:tsne}
\begin{figure}[htb]
    \begin{center}
        \includegraphics[width=0.6\linewidth]{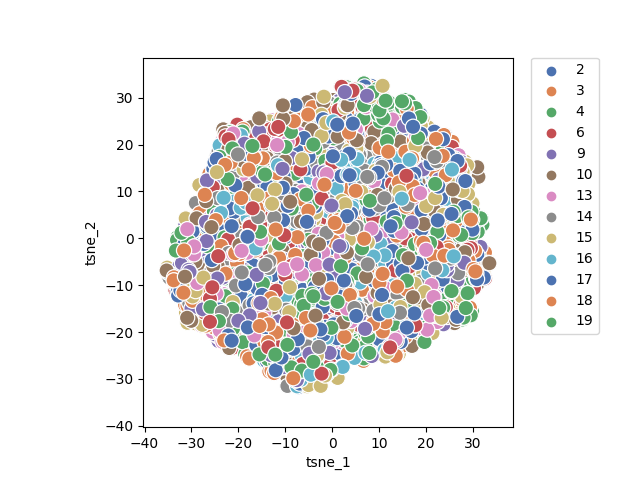}
    \end{center}
    \caption{
    \textbf{Visualisation of the specialised parameters for experiment 2 --} 
	The fully-connected layers parameters, after adaptation on the context set, are embedded with
	t-SNE and labelled by object. This is done for each sampled task of the largest test set (13
	objects, 3814 tasks). The absence of clusters indicates that the specialised parameters do not
	encode object-specific information on the hand-object pose.} \label{fig:tsne}
\end{figure}

\begin{figure}[hb]
    \centering
    \begin{subfigure}[t]{0.49\textwidth}
        \centering
        \includegraphics[width=\textwidth]{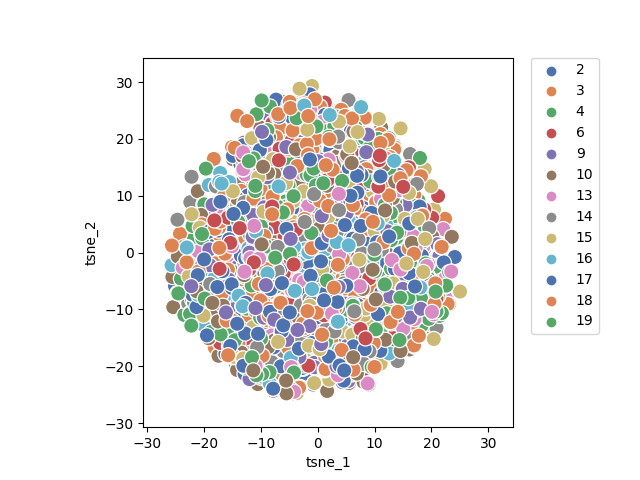}
        \caption{\textbf{Weights of first layer: experiment 1} -- Hand pose only.}
    \end{subfigure}
    \hfill
    \begin{subfigure}[t]{0.49\textwidth}
        \centering
        \includegraphics[width=\textwidth]{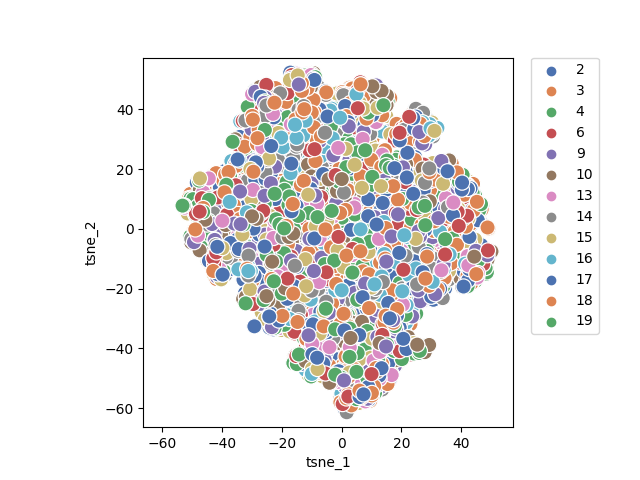}
        \caption{\textbf{Weights of second layer: experiment 1} -- Hand pose only.}
    \end{subfigure}
    \hfill
    \begin{subfigure}[t]{0.49\textwidth}
        \centering
        \includegraphics[width=\textwidth]{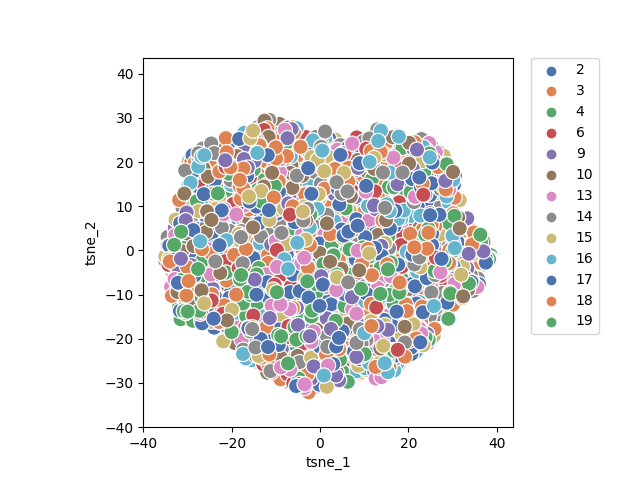}
        \caption{\textbf{Weights of first layer: experiment 2} -- Joint hand-object pose.}
    \end{subfigure}
    \hfill
    \begin{subfigure}[t]{0.49\textwidth}
        \centering
        \includegraphics[width=\textwidth]{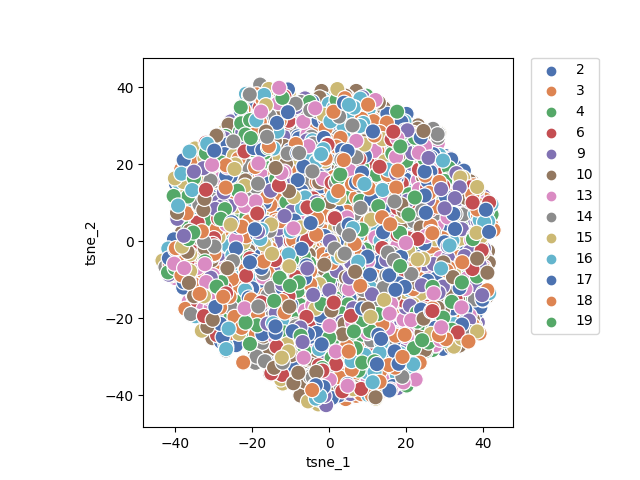}
        \caption{\textbf{Weights of second layer: experiment 2} -- Joint hand-object pose.}
    \end{subfigure}
    \caption{\textbf{Visualisation of weights per layer} -- The t-SNE embeddings of the adapted weights of each layer, for both experiments,
    show no signs of separate clusters for the 13 different objects. This means that they mostly encode the same information regarding the
    hand or joint hand-object pose for all objects.}
    \label{fig:tsne_weights}
\end{figure}

\begin{figure}[tbp]
    \centering
    \hfill
    \begin{subfigure}[t]{0.49\textwidth}
        \centering
        \includegraphics[width=\textwidth]{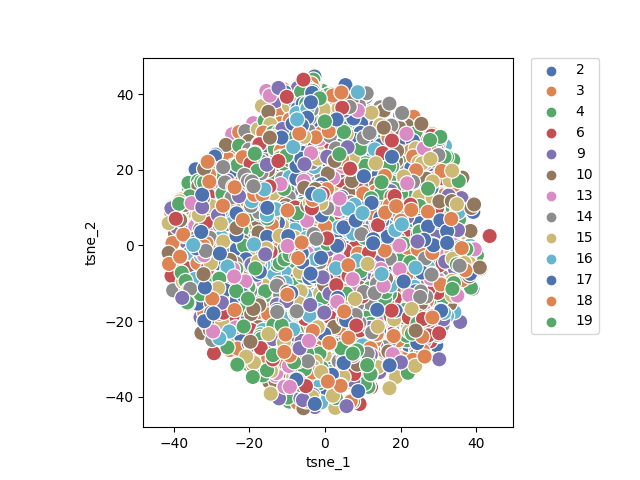}
        \caption{\textbf{Biases of first layer: experiment 1} -- Hand pose only.}
    \end{subfigure}
    \hfill
    \begin{subfigure}[t]{0.49\textwidth}
        \centering
        \includegraphics[width=\textwidth]{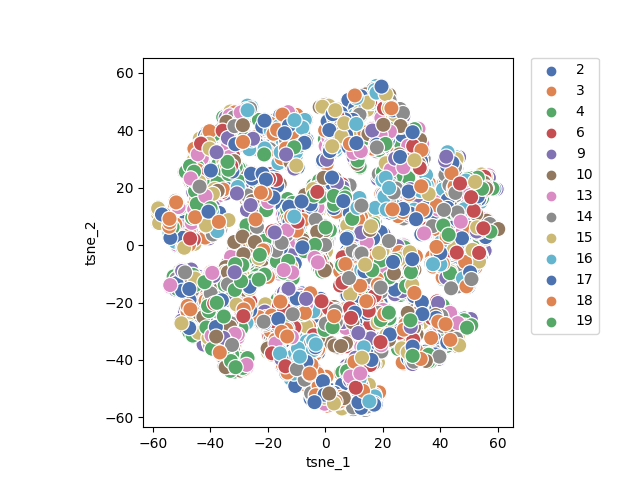}
        \caption{\textbf{Biases of second layer: experiment 1} -- Hand pose only.}
    \end{subfigure}
     \hfill
    \begin{subfigure}[t]{0.49\textwidth}
        \centering
        \includegraphics[width=\textwidth]{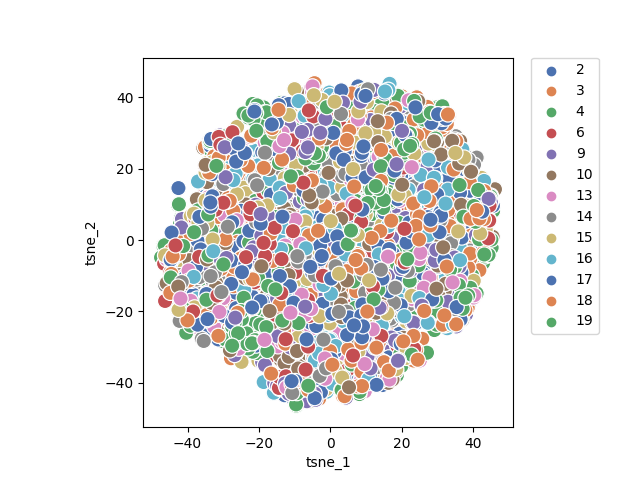}
        \caption{\textbf{Biases of first layer: experiment 2} -- Joint hand-object pose.}
    \end{subfigure}
    \hfill
    \begin{subfigure}[t]{0.49\textwidth}
        \centering
        \includegraphics[width=\textwidth]{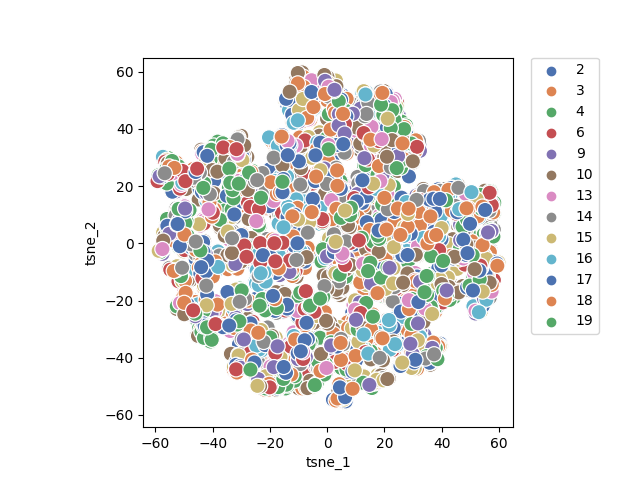}
        \caption{\textbf{Biases of second layer: experiment 2} -- Joint hand-object pose.}
    \end{subfigure}
    \caption{\textbf{Visualisation of biases per layer} -- The t-SNE embeddings of the adapted biases of each layer, for both experiments,
    show that no structure is present in the first layer, while they appear to be clustered in some way for the last layer,
    the final 3D regression layer. However, those clusters do not coincide with the objects themselves, therefore the biases of the last
    layer may encode information specific to other aspects of the image features.}
    \label{fig:tsne_biases}
\end{figure}

We plotted the t-SNE embeddings of all parameters of the network's head, after adaptation on the context set, in \cref{fig:tsne}. The
absence of clusters reveals that no object-specific information is encoded in the parameters during adaptation.
In \cref{fig:tsne_weights}, the weights alone of each layer of the network's head are embedded,
and the biases in \cref{fig:tsne_biases}.
Only the biases of the final layer seem to reveal some structure after adaptation, although it is
unknown what this corresponds to.

\subsection{Interpretation of the gradients norm during adaptation}\label{appendix:gradnorm}
We here provide evidence to assess the hypothesis that the adaptation phase produces
smaller gradients for the object keypoints than for the hand keypoints. In \cref{sec:analysis},
we make this postulate to conclude that the hand keypoint regression task should not be severely impacted.

\begin{table}[h]
    \centering
    \caption{\textbf{Average gradient norm per adaptation step: exp. 1 vs exp.} --
    Entries follow an \textit{exp1/exp2} format, with larger values in bold. \\}
        \begin{tabular}{ cccccc }
            \addlinespace
            \toprule
             Step & Tomato soup can & Banana & Scissors & Foam brick & Mug\\
            \midrule
             1  & 204/\textbf{394} & 231/\textbf{638} & 270/\textbf{677} & 195/\textbf{380} & 236/\textbf{505}  \\
             2  & 193/\textbf{360} & 219/\textbf{576} & 255/\textbf{613} & 185/\textbf{338} & 225/\textbf{458}  \\
             3  & 182/\textbf{330} & 207/\textbf{522} & 242/\textbf{557} & 176/\textbf{302} & 214/\textbf{416}  \\
             4  & 173/\textbf{303} & 197/\textbf{475} & 229/\textbf{509} & 167/\textbf{273} & 205/\textbf{381}  \\
             5  & 164/\textbf{280} & 187/\textbf{435} & 217/\textbf{467} & 159/\textbf{248} & 195/\textbf{349}  \\
             6  & 155/\textbf{259} & 178/\textbf{401} & 207/\textbf{430} & 152/\textbf{227} & 187/\textbf{322}  \\
             7  & 148/\textbf{241} & 169/\textbf{371} & 197/\textbf{398} & 145/\textbf{209} & 179/\textbf{298}  \\
             8  & 141/\textbf{225} & 161/\textbf{344} & 188/\textbf{369} & 139/\textbf{194} & 172/\textbf{277}  \\
             9  & 134/\textbf{210} & 153/\textbf{321} & 180/\textbf{344} & 133/\textbf{181} & 165/\textbf{259}  \\
             10 & 128/\textbf{198} & 146/\textbf{301} & 172/\textbf{322} & 127/\textbf{169} & 159/\textbf{243}  \\
            \bottomrule
        \end{tabular}
    \label{tab:grad_norms}
\end{table}

In \cref{tab:grad_norms} we show the norm of the gradients during the adaptation phase.
For both experiments, on the same randomly sampled test objects, the gradient norms of the first 10
adaptation steps are averaged over batches of the same task. 
Both models were trained on the same 6 objects. Experiment 2 has consistently much larger gradients,
meaning that jointly regressing the hand and object poses requires deviating further away from the
initialisation during adaptation.  This could reflect a poor initialisation of meta-parameters.


\end{document}